\documentclass{article}
\usepackage{spconf,amsmath,graphicx}
\usepackage{algorithm}
\usepackage{algpseudocode}
\usepackage{amsmath}
\usepackage{graphics}
\usepackage{epsfig}
\usepackage[colorlinks, linkcolor=green, anchorcolor=green, citecolor=green,]{hyperref}

\title{DAF-net: a saliency based weakly supervised method of dual attention fusion for fine-grained image classification}
\name{ZiChao Dong\textsuperscript{1}, JiLong Wu\textsuperscript{2}, TingTing Ren\textsuperscript{2}, Yue Wang\textsuperscript{3}, MengYing Ge\textsuperscript{4}}
\address{The Chinese University of Hong Kong\textsuperscript{1},Geng Mei AI lab\textsuperscript{[2, 4]}, The University of Hong Kong\textsuperscript{3}.\\
1155129705@link.cuhk.edu.hk,wujilong@igengmei.com}
\begin{document}
%
\maketitle
\begin{abstract}
Fine-grained image classification is a challenging problem, since the difficulty of finding discriminative features.To handle this circumstance,basically,there are two ways to go.One is use attention based method to focus on informative areas,while the other one aims to find high order order between features.Further,for attention based method there are two directions,activation based and detection based,which are proved effective by scholars.However,rare work focus on fusing two types of attention with high order feature. 

In this paper,we propose a novel 'Dual Attention Fusion'(DAF) method which fuse two types of attention and use them to as 'PAF'(part attention filter) in deep bilinear transformation module to mine the relationship between separate parts of an object.Briefly,our network constructed by a student net who attempt to output two attention maps and a teacher net uses these two maps as empirical information to refine the result. The experiment result shows that only student net could get 87.6\% accuaracy in CUB dataset while cooperating with teacher net could achieve 89.1\% accuarcy. 

\end{abstract}
\section{Introduction}
\label{sec:intro}
Fine-grained image classification, also known as sub-category recognition, is a hot research topic in computer vision and pattern recognition in recent years. The purpose of fine-grained classification is to subclass the images belonging to the same basic category (cars, dogs, flowers, birds, etc.) in a more detailed way. However, due to the subtle inter-class differences and large intra-class differences among sub-categories, the classification of fine-grained images is more difficult than ordinary image classification tasks. Deep learning has brought tremendous advances to many areas of computer vision \cite{wy1, wy2, wy3}, however, its performance in fine-grained classification is not satisfactory on account of the difficulty of finding informative regions and extracting discriminative features. The result of classification is even worse due to a variety of different backgrounds, poses and shooting Angles\cite{wy4}.
Previous research\cite{wy5, wy6, wy7, wy8, wy9, wy10, wy11} of fine-grained classification rely on human annotations. However, it is so expensive to annotate fine-grained features that these approaches are impractical. Some improvements \cite{wy12, wy13, wy14, wy15} employed unsupervised learning have been proposed, which can localize informative regions automatically. Unfortunately, most of the work had low accuracy. NTS-net innovatively put forward a multi-agent cooperative learning mechanism which have Navigator, Teacher and Scrutinizer that can increase accuracy. However, we found it is hard to accurately detect separate bounding boxes, such as a single head.

In this paper, we combine fine-grained image classification with saliency detection to attach more attention on the small and distinctive parts. Since saliency detection is a segmentation task with pixel-level annotation, we use a weakly supervised box-level segmentation map instead. In many detection-based works, we found it is hard to directly locate the key areas in an object. In order to solve this problem, we make it two stages, using activation based heatmaps to locate the instance and saliency segmentation to accurately detect informative parts on an instance. We also found the previous work; especially fancy attention mechanism is not that satisfying and not that explainable. we designed an unambiguous network to show users whether it is concentrating on the parts that users wish to.

Our main contributions can be summarized as follows: 
\begin{itemize}
    \item The first to introduce saliency detection to fine-grained image classification task by adding SPPN module.
    \item Design a location heatmap generation mechanism to boost SPPN(saliency part proposal network) module.
    \item Design R\_IOU loss and R\_area loss to get separate and accurate saliency part proposals, which constructed as box-level saliency segmentation mask.
    \item Use a self-supervised mechanism to refine saliency proposals.
    \item Design a simple and effective method to fuse dual attention which is extremely helpful for deep bilinear transformation module as PAF(part attention filter).
    \item Use a knowledge distillation based method to boost student and teacher net.
\end{itemize}

\section{RELATED WORK}
\label{sec:format}

\subsection{Fine-grained image classification}
\label{ssec:subhead}
The difficulty of fine-grained image classification is subtle inter-class differences and large intra-class differences among sub-categories. In order to successfully carry out fine-grained classification of two very similar species, the most important thing is to find the informative region that can distinguish the two species in the image. 

Spatial Transformer Network\cite{wy18} achieve spatial invariability by predicting the location of informative regions and then correcting the image to an ideal position. Bilinear CNN models\cite{wy19} uses two feature extractors, which coordinate with each other to detect informative regions and extract features. \cite{wy12} propose an automatic component detection method for fine image recognition, learning part detectors and part saliency maps. RA-CNN \cite{wy20} uses mutually reinforcing methods to recursively learn discriminative region attention and region-based feature representation. DVAN \cite{wy13} improves the diversity of visual attention to extract the maximum discriminant features. It consists of four parts: attention region generation, CNN feature extraction, diverse visual attention and classification. FCN attention\cite{wy21} is a full-convolution attentional network based on reinforcement learning, which can adaptively select multi-task-driven attention regions. Because it is based on FCN architecture, it is more efficient and can locate multiple object parts and extract features of multiple attention regions at the same time. Short-term Memory network (LSTM) is unified into a deep loop architecture called HSnet\cite{wy22}. Therefore, HSnet(1) produces the proposal of image parts with information, and (2) integrates all the proposals for final-grained recognition.

\subsection{Object detection}
\label{ssec:subhead}

Early object detection methods mainly use SIFT \cite{wy23} or HOG \cite{wy24} features. Recently, deep learning-based object detection methods have shown dramatic improvements. Two-stage approaches like R-CNN \cite{wy25}, OverFeat \cite{wy26} and SPPnet \cite{wy27} adopt traditional image-processing methods to generate object proposals and perform category classification and bounding box regression. Faster R-CNN \cite{wy28} first proposes a unified end-to-end framework for object detection. It introduces a region proposal network (RPN) which shares the same backbone network with the detection network to replace the original standalone time-consuming region proposal methods. On the other hand, one-stage methods which are popularized by YOLO \cite{wy29, wy30, wy31} and SSD \cite{wy32} improve detection speed over Faster R-CNN \cite{wy28} by employing a single-shot architecture. RetinaNet \cite{wy16} proposes a new focal loss to address the extreme foreground-background class imbalance which stands out as a central issue in one-stage detectors. Feature Pyramid Networks (FPN) \cite{wy33} focuses on better handling multi-scale objects and generates anchors from multiple feature maps.

\subsection{Saliency Detection}
\label{ssec:subhead}
Visual saliency prediction in computer vision aims at estimating the locations in an image that attract the attention of humans. Salient points in an image are understood as a result of a bottom-up process where a human observer explores the image for a few seconds with no particular task in mind. These data are traditionally measured with eye-trackers, and more recently with mouse clicks \cite{d1} or webcams \cite{d3}. The salient points over the image are usually aggregated and represented in a saliency map, a single channel image obtained by convolving each point with a Gaussian kernel. As a result, a gray-scale image or heatmap is generated to represent the probability of each corresponding pixel in the image to capture the human attention\cite{d4}. These saliency maps have been used as soft-attention guides for other computer vision tasks, and also directly for user studies in fields like marketing.

However,it is expensive to acquire pixel-level annotations of saliency map as ground truth.Instead, we try to use a weakly supervised method by only use image level classification ground truth to guide us. In \cite{d11}, they try to combine object detection with fine-grained image classification to get higher accuracy. Early works \cite{d2} attempt to use image-level annotations to get pixel-level segmentation results. In our work, instead of using fully connected convolutional neural networks to get segmentations mask\cite{d5}, we use feature proposal network to do part-level object detection which could provide us box-level saliency information inside an object. We hope to get distinctive features from those saliency areas. Like \cite{d6,d7,d7,d8}, we use bonding box information to get segmentation masks. Further, inspired by \cite{d10}, we found it is feasible to concatenate saliency detection map with image as a attention guiding. 

\subsection{Activation Based Heatmap}
\label{ssec:subhead}
Recently, several heatmap methods sourcing from the activation layer of backbone have been proposed to boost the fine-grained image classification. Typically, SCDA\cite{F1} is a kind of unsupervised method  to locate the main region in the fine-grained images, without the help of image label or bounding box annotation. It takes the activation-map after convolution layer , ReLU layer or pooling layer and then add up the activation through the deep direction ,calculates the mean of the activation-map. As for the value in the activation-map who is higher than the mean value, they will keep unchanged while others will be erased to zero. In this way ,the SCDA\cite{F1} method generate a activation-map which could be coarsely considered as the object segmentation map. Apart from that, it is also an elegant way to guide the following network as an attention.

\subsection{Bilinear Pooling}
\label{ssec:subhead}

Bilinear pooling\cite{dd1} is proposed to obtain invariant and discriminative global representation for convolutional feature, which achieved the state-of-the-art results in many fine-grained datasets. However, the high-dimensionality issue is caused by calculating pairwise interaction between channels, thus dimension reduction methods are proposed. Specifically, low-rank bilinear pooling\cite{dd3} proposed to reduce feature dimensions before conducting bilinear transformation, and compact bilinear pooling\cite{dd2} proposed a sampling based approximation method, which can reduce feature dimensions by two orders of magnitude without performance drop. Moreover, feature matrix normalization\cite{dd4} is proved to be important for bilinear feature, while we do not use such technics in our deep bilinear transformation since calculating such root is expensive and not practical to be deeply stacked in CNNs. Second-order pooling convolutional networks\cite{dd5} also proposed to integrate bilinear interactions into convolutional blocks, while they only use such bilinear features for weighting convlutional channels. In \cite{dd6}, they manage to use semantic grouping to select relative features and combine them together for further bilinear feature transformation.

\subsection{Knowledge Distillation}
\label{ssec:subhead}
Knowledge distillation with neural networks was pioneered by Hinton \cite{K1} and Bucila \cite{K2}, which is transfer learning method that aims to improve training of student work by relying on knowledge borrowed from a powerful teacher network. The framework could compress an ensemble of deep networks(teacher) into a student network of similar depth \cite{K3}. To do so, the student is trained to predict the output of the teacher as well as the true classification labels. 

However, all previous works related on this method focus on the theme about model compression\cite{K4}. In this paper, we are the first one to introduce knowledge distillation method to fine-grained classification while reduce the distinguish of parameters distribution between the teacher net and student net in the training process. 

\section{Method}
\label{sec:pagestyle}

\subsection{Method Overview}
\label{ssec:subhead}

Our method consists of two parts, student net and teacher net respectively.As for student net, received an image as input, a backbone serves as a feature extractor.Similar to retina-net \cite{d12}, we add a feature pyramid network to predict the bounding box in multi-level.Inspired by \cite{d13}, we use the activation map of the three layers in last three blocks of backbone to get the location-heatmap in the mean time.The location-heatmap is considered as a coarse location for the object, which has two effects:1. A judgement for foreground and background for the object detection task. 2. A coarse attention for teacher net to locate the object. Further, the feature pyramid network is also used to operate the region proposal network, where the saliency part of an object could be detected by a bounding box. In order to get the accurate informative parts, it is reasonable that the bounding boxes should be approximately 1/N size of the total area of the object and the bounding boxes should have rare intersections since we want each bounding box matches a distinctive part. To manage that, we designed two losses, R\_area loss and R\_IOU loss respectively. Since the detection is lack of ground truth, we operate a self-supervise method by using a ranking loss. 

After student net,a location-heatmap map and a saliency map constructed by the bounding boxes are acquired.We concatenate the two attention maps with original image and feed it in teacher net.Firstly, the two attention maps hint the location and saliency part of the object.Secondly,we use saliency part mask to filter the feature map.Precisely,discriminative feature is highlighted,which is a great help for semantic grouping and group bilinear part.After the deep bilinear transformation,the final feature vector is obtained.Since teacher net outperform student net,we use distillation and hint learning method to encourage student net learn better.The final result for infence is an ensemble between results of student net and teacher net.

\begin{algorithm}
\caption{ Framework of Our System.} 
\label{alg:Framwork} 
\begin{algorithmic}[1] 
\Require 
The origin image
\Ensure 
the classification result of the origin image; 
\State 
Get the attention map, detection map and first stage prediction; 
\label{code:fram:extract} 
\State 
Concatenate the attention maps and origin pic as the input of the teacher net;
\label{code:fram:trainbase} 
\State 
Construct the distillation structure between student net and teacher net;
\label{code:fram:add} 
\State 
Get the prediction from the teacher net;
\label{code:fram:classify} 
\State 
ensemble the result between teacher net and student net; 
\label{code:fram:select} \\ 
\Return $E_n$; 
\end{algorithmic} 
\end{algorithm}

\begin{figure}[htb]

\begin{minipage}[b]{1.0\linewidth}
  \centering
  \centerline{\includegraphics[width=8.5cm]{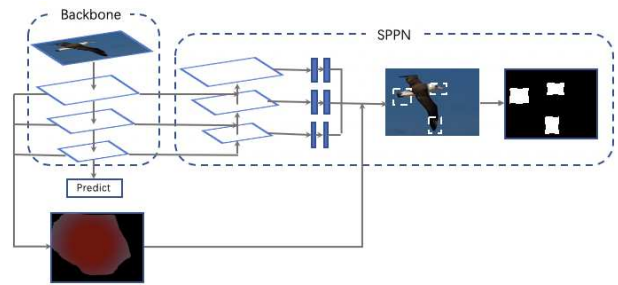}}
\end{minipage}
\caption{Pipeline for SPPN}
\label{fig:res}
\end{figure}

\subsection{Location Heatmap Module}
\label{ssec:subhead}
In our method, we use location heatmap,similar to the SCDA\cite{F1} method, to separate the foreground and background.Three ReLU layers corresponding to last three blocks of backbone in the student network are aggregated to generate the heatmap.For each, we add up the activation tensors through the depth direction.Thus, the hwd 3D tensor becomes an hw 2D tensor, which we call the “aggregation map” and name it A, we will use A in the following.Then, we calculate the mean value of all the positions in A as the threshold to decide which positions localize objects. Consequently, the position whose activation response is higher than the mean value indicates the main object, we set it to one, and other positions zero.We concatenate the three “aggregation map” together, and normalize it.In this way, we get the location heatmap which involves not only the low dim texture information but also the high dimension whole target shape.The location heatmap will be used in two places. First, it is put into SPPN network, as it aggregates the object region, it can be used as the coarse target location, which directs the SPPN network in regressing the location of the proposals.Second, in the teacher net, the location heatmap will be used as an input information together with the saliency map.

\subsection{SPPN Module}
\label{ssec:subhead}

Saliency part proposal network is a subnet cascaded behind our backbone, in order to predict the saliency parts of an object. Like widely used region proposal network(RPN) in object detection, our SPPN aims to yield saliency part proposals. Benefit from location-heatmap, we can select the positive samples who falls in the activated area in that heatmap. For others, we consider them as negative samples and neglect the following classification and regression for them. By doing so, the difficulty for SPPN to learn is lower. 

Similar to retina-net, classification head and regression head are added in all three level of feature map to predict the convincing score and saliency bounding box corresponding to different size of target. We choose top-N informative regions after NMS and use a self-supervised method to further boost SPPN. NTS-net uses a similar way, however they only remain the information inside the bounding box and remove others. As far as our concerned, it could be considered as high-level crop along with losing the context messages. We would argue that it is more reasonable to restore the information inside the box while keeping the context message in the meantime. 

In our method, we designed a fancy method to for evaluate the bounding boxes. For the original input image and every bounding box chosen from N boxes, firstly we keep the pixel in the box unchanged. Secondly, we add noise in the corresponding area for other N-1 boxes. Thirdly, we operate heavy Gaussian blur in other areas. N processed images are acquired by the method and we feed them into our backbone to get N scores. A ranking loss is implemented in order to realize the self-supervise mechanism.

\begin{figure}[htb]

\begin{minipage}[b]{1.0\linewidth}
  \centering
  \centerline{\includegraphics[width=8.5cm]{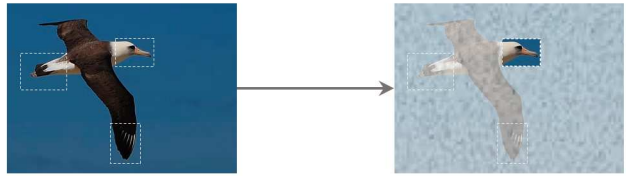}}
\end{minipage}
\caption{Illustration of the prepossessing of each saliency part proposal.The output will be fed into backbone again to evaluate the value of the proposal.}
\label{fig:res}
\end{figure}

\subsection{PAF Module}
\label{ssec:subhead}
PAF, stands for part attention filter, is designed to highlight the discriminative parts of an object.Firstly,saliency part mask is input in adaptive layer,which out put a part attention mask with the same shape as the corresponding feature map.Secondly,it is used as a filter to highlight informative feature while ignore useless areas.Inspired by\cite{dd6},we use semantic grouping to gather feature map who focus on similar areas and group bilinear to calculate and aggregate intra-group pairwise interactions.The pipeline is shown below:

\begin{figure}[htb]

\begin{minipage}[b]{1.0\linewidth}
  \centering
  \centerline{\includegraphics[width=8.5cm]{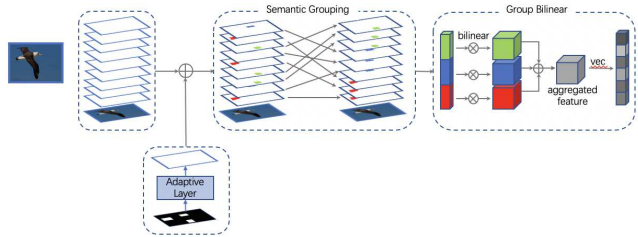}}
\end{minipage}
\caption{We use PAF module to guide the semantic grouping.}
\label{fig:res}
\end{figure}

\subsection{Distillation Framework}
\label{ssec:subhead}

Distillation framework is applied to improve the similar of parameters distribution between teacher net and student net which is borrowed from knowledge distillation structure\cite{P1}.  In this paper, we introduce intermediate-level hints from the teacher hidden layers to guide the training process of the student\cite{P2}. The distillation framework can be summarized as follows.
First, we introduce the hints which are defined as the output of teacher net hidden layers.
Second, the guided layer is selected from the student net layers to learn from the teachers hint layer.
Third, we add a regressor whose output match the size of hint layer to the guide layer.
Finally, the soft target cross entropy loss is applied on the head of the distillation framework. 
Then the loss function of distillation framework is defined as 
\begin{equation}
\begin{aligned}
L_{HT}(F_{hint}, F_{guide}) = \frac{\alpha}{2}||F_{hint}-r(F_{guide})||^2\\
 +\beta L_c(L_{soft}, L_{target})
\end{aligned}
\end{equation}

where $L_c$ is the soft target cross entropy loss.

\subsection{CSF Module}
\label{ssec:subhead}

CSF stands for class score fusion,which is used to ensumble results from student net and teacher net.In \cite{dd7},authors use a module to fuse two scores from two streams in video classification task.Similarly,our teacher-student net architecture can also be considered as two-stream module.So we designed a class score fusion module to fuse the result vector from two nets by using fully connected layer.Cross entropy loss is then used to optimise this module.

\begin{figure}[htb]

\begin{minipage}[b]{1.0\linewidth}
  \centering
  \centerline{\includegraphics[width=8.5cm]{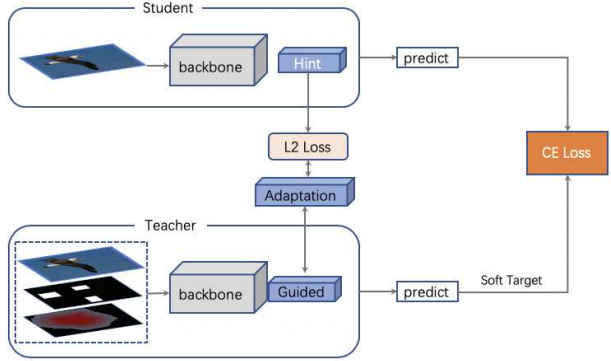}}
\end{minipage}
\caption{Whole picture of our framework.}
\label{fig:res}
\end{figure}




\subsection{Loss function and Optimization}
\label{ssec:subhead}

In our model, we have designed several loss functions in order to get better accuracy. They are illustrated as follow.

\subsubsection{R\_IOU Loss}
\label{ssec:subsubhead}

In order to get precise discriminating parts of an object, we attempt to make our bounding boxes have little intersection with each other. otherwise, more than one bounding box will predicted corresponding to same part. Inspired by dice loss, we simply add a hype-parameter alpha to control the penalty of the intersection. The formula is shown below:

\begin{equation}
\begin{aligned}
L_{IOU}(B_1, B_2) = \alpha\frac{|B_1\bigcap B_2|}{||B_1|+|B_2||}
\end{aligned}
\end{equation}

\subsubsection{R\_Area Loss}
\label{ssec:subsubhead}

In our experiment, we find that only use R\_IOU loss is not enough to make bounding boxes accurately locate single parts of an object since the box are often big and box another areas rather the distinctive part. So we would argue that another restrict to reduce the area of bounding box is required. Thanks to location-heatmap, we can coarsely estimate the area S of the total object. Then we set a hype-parameter N indicated that how many parts are there in a object. The number of parts N could be changed due to different type of objects. For instance, in the bird dataset CUB-200-2011, we set N=4. The formula is shown below:



\subsubsection{Distillation Loss}
\label{ssec:subsubhead}

In order to let student and teacher net collaborate better, we design a distillation loss which combine two parts, hint learning and soft target learning. The formula is shown below:

\begin{equation}
\begin{aligned}
L_{HT}(F_{hint}, F_{guide}) = \frac{\alpha}{2}||F_{hint}-r(F_{guide})||^2\\
 +\beta L_c(L_{soft}, L_{target})
\end{aligned}
\end{equation}

\subsubsection{Concentrate Loss}
\label{ssec:subsubhead}
For total M saliency part proposals, we also introduced a concentrate loss to combine their effects. Technically,each box, like we mentioned above, could be convert to an unique image and we feed these M images into our student net(without SPPN part) again to get M logits. Each logit corresponding to the probability for each class. We use an ensemble method to get the fusion logit of M boxes. In our experiment, we directly use average of the M logits. The formula is shown below:

\begin{equation}
\begin{aligned}
L_{con}(F_logit, GT) = Cross_Entropy(F_logit,GT)
\end{aligned}
\end{equation}

\subsubsection{Ranking Loss}
\label{ssec:subsubhead}

Similar to what they do in NTS-net, we use ranking loss to penalize pairs with wrong orders.Let X = {X1, X2, · · · , Xn} denote the objects to rank, and Y = {Y1, Y2, · · · , Yn} the indexing of the objects, where $Yi \geq Yj$ means Xi should be ranked before Xj. We use pair-wise ranking approach,  Suppose F(Xi,Xj) only takes a value from {1,0}, where F(Xi,Xj) = 0 means Xi is ranked before Xj. Then the loss is defined on all pairs, and the goal is to find an optimal F to minimize the average number of pairs with wrong order. The formula is shown below:

\begin{equation}
\begin{aligned}
L_{Ranking}(F, X, Y) = \sum_{(i, j):Y_i<Y_j}F(x_i, x_j)
\end{aligned}
\end{equation}

\subsubsection{Total Loss for DAF-net}
\label{ssec:subsubhead}

The total loss constructed by three parts: student net,teacher net and the class score fusion(CSF) module.

\begin{equation}
\begin{aligned}
L_{Total} = \alpha L_{student} + \beta L_{teacher} + \gamma L_{CSF}
\end{aligned}
\end{equation}

For loss of student net, there are five parts. Prediction loss indicated the cross entropy loss.

\begin{equation}
\begin{aligned}
L_{Student} = L_{Ranking} + L_{Area} + L_{IOU} + L_{HT} + L_{Pred} + L_{Con}
\end{aligned}
\end{equation}

For loss of teacher net and CSF module, only one loss are used separately. 

\begin{equation}
\begin{aligned}
L_{Teacher} = L_{Pred}
\end{aligned}
\end{equation}

\begin{equation}
\begin{aligned}
L_{CSF} = L_{Pred}
\end{aligned}
\end{equation}

\section{Experiments}
\label{sec:copyright}
\subsection{Dataset}
\label{sssec:subsubhead}
We comprehensively evaluate our algorithm on Caltech-UCSD Birds (CUB-200-2011) [42], Stanford Cars [23] and FGVC Aircraft [35] datasets, which are widely used benchmark for fine-grained image classification. We do not use any bounding box/part annotations in all our experiments. Statistics of all 3 datasets are shown in Table. 1, and we follow the same train/test splits as in the table.\\
{\bfseries Caltech-UCSD Birds.} CUB-200-2011 is a bird classification task with 11,788 images from 200 wild bird species. The ratio of train data and test data is roughly 1 : 1. It is generally considered one of the most competitive datasets since each species has only 30 images for training.\\
{\bfseries Stanford Cars.} Stanford Cars dataset contains 16,185 images over 196 classes, and each class has a roughly 50-50 split. The cars in the images are taken from many angles, and the classes are typically at the level of production year and model (e.g. 2012 Tesla Model S).\\
{\bfseries FGVC Aircraft.} FGVC Aircraft dataset contains 10,000 images over 100 classes, and the train/test set split ratio is around 2 : 1. Most images in this dataset are airplanes. And the dataset is organized in a four-level hierarchy, from finer to coarser: Model, Variant, Family, Manufacturer.

\begin{table}[h] 
\begin{tabular}{p{3cm}|p{1cm}|p{1cm}|p{1cm}}
\hline
\hline
Dataset & Class & Train & Test \\ 
\hline 
CUB-200-2011 & 200& 5,994& 5,794\\
\hline
Stanford Cars & 196& 8,144& 8,041\\
\hline
FGVC Aircraft & 100& 6,667& 3, 333\\
\hline
\hline
\end{tabular}
\caption{Statistics of benchmark datasets}
\end{table}

\subsection{Implementation Details}
\label{sssec:subsubhead}

In all our experiments, we preprocess images to size 448, and we fix M=3
which means 3 regions are used to train student network for each image (there
is no restriction on hyper-parameter). We use fully-convolutional network ResNet-50 as feature extractor and use Batch Normalization as regularizer. We use Momentum SGD with initial learning rate 0.01 and multiplied by 0.5 after 50 epochs, and we use weight decay 1e-4\\. The NMS threshold is set to 0.25. Our model is robust to the selection of hyper-parameters. 

\subsection{Quantitative Results}
\label{sssec:subsubhead}
Overall, our proposed system outperforms all previous methods expect for WS-DAN. Since we do not use any bounding box/part annotations, we do not compare with methods which depend on those annotations. Table. 2 shows the comparison between our results
and previous best results in CUB-200-2011. Both Inception-V3 and ResNet-50 are strong baselines. However, we found that combining with activation based heatmap, ResNet-50 is a better choice. So in our framework, we choose ResNet-50 as backbone in both teacher and student net and achieve 89.1\% accuracy.
\begin{table}[h] 
\begin{tabular}{p{5cm}|p{3cm}}
\hline
\hline
Method &  top-1 accuracy \\ 
\hline 
VGG19&77.8\%\\
\hline
ResNet-50&83.5\%\\
\hline
Inception-V3&83.7\%\\
\hline
\hline
Bilinear-CNN & 84.1\%\\
\hline 
ST-CNN& 84.1\%\\
\hline
FCAN& 84.3\%\\
\hline
ResNet-50&84.5\%\\
\hline 
PDFR& 84.5\%\\
\hline 
RA-CNN&85.3\%\\
\hline 
HIHCA&85.3\%\\
\hline 
Boost-CNN&85.6\%\\
\hline 
DT-RAM&86.0\%\\
\hline 
MA-CNN&86.5\%\\
\hline
DFL-CNN&87.4\%\\
\hline 
NTS-net&87.5\%\\
\hline
TASN&87.6\%\\
\hline
MPN-COV&88.7\%\\
\hline
WS-DAN&89.4\%\\
\hline
\hline
\textbf{Our student net}&\textbf{87.6}\%\\
\hline 
\textbf{Our student+teacher net}&\textbf{89.1}\%\\
\hline 
\hline

\end{tabular}
\caption{Experimental results in CUB-200-2011}
\end{table}

\subsection{Ablation Study}
\label{sssec:subsubhead}

In order to analyze the influence of different components in our framework, we design different runs in CUB-200-2011 and report the results in Table. 3. The experiments are focusing on our separate contributions.

\begin{table}[h] 
\begin{tabular}{p{5cm}|p{3cm}}
\hline
\hline
Method & top-1 accuracy  \\ 
\hline 
ResNet-50 baseline & 83.5\%\\
\hline 
\hline
ResNet-50+SPPN&87.4\%\\
\hline
ResNet-50+SPPN+LHM&87.6\%\\
\hline
ResNet-50+SPPN+LHM+teacher & 88.9\%\\
\hline
\textbf{Complete DAF-net(+distillation)}& \textbf{89.1}\%\\
\hline 
\hline 

\end{tabular}
\caption{Study of influence factor in CUB-200-2011}
\end{table}


\bibliographystyle{IEEEbib}
\bibliography{refs}

\end{document}